\documentclass[conference]{IEEEtran}
\IEEEoverridecommandlockouts
\usepackage{cite}
\usepackage{amsmath,amssymb,amsfonts}
\usepackage{algorithmic}
\usepackage{graphicx}
\usepackage[margin=0.75in]{geometry}
\usepackage{textcomp}
\usepackage[caption=false]{subfig}
\usepackage{float}
\newcommand{\squeezeup}{\vspace{-3.5mm}}
\def\BibTeX{{\rm B\kern-.05em{\sc i\kern-.025em b}\kern-.08em
    T\kern-.1667em\lower.7ex\hbox{E}\kern-.125emX}}
\begin{document}

\title{Classification of Building Information Model (BIM) Structures with Deep Learning
}


\author{\IEEEauthorblockN{Francesco Lomio \IEEEauthorrefmark{1}, 
Ricardo Farinha \IEEEauthorrefmark{2}, 
Mauri Laasonen \IEEEauthorrefmark{3}, 
Heikki Huttunen \IEEEauthorrefmark{1}}
\IEEEauthorblockA{\IEEEauthorrefmark{1}Tampere University of Technology, Tampere, Finland\\}
\IEEEauthorblockA{\IEEEauthorrefmark{2}Sweco Finland Ltd., Tampere, Finland\\}
\IEEEauthorblockA{\IEEEauthorrefmark{3}Sweco Structures Ltd., Tampere, Finland\\
Email: francesco.lomio@tut.fi}
}

\maketitle

\begin{abstract}
In this work we study an application of machine learning to the construction industry and we use classical and modern machine learning methods to categorize images of building designs into three classes: Apartment building, Industrial building or Other. No real images are used, but only images extracted from Building Information Model (BIM) software, as these are used by the construction industry to store building designs. For this task, we compared four different methods: the first is based on classical machine learning, where Histogram of Oriented Gradients (HOG) was used for feature extraction and a Support Vector Machine (SVM) for classification; the other three methods are based on deep learning, covering common pre-trained networks as well as ones designed from scratch. To validate the  accuracy of the models, a database of 240 images was used. The accuracy achieved is 57$\%$ for the HOG + SVM model, and above 89$\%$ for the neural networks.
\end{abstract}


\section{Introduction}
In the past decades, machine learning, and more recently  deep learning, has found applications in a big variety of fields, from data analysis to computer vision and image recognition. Thanks to the advances in this area, their applications have already contributed to revolutionize many industries.

Particularly, the field of image recognition has seen an increase in development in the recent years. In the automotive industry, for example, the use of deep learning algorithm has allowed self-driving cars to recognize lanes and other obstacles without the need for more expensive and complex tools \cite{HuvalAnDriving}. But the range of applications extends also to fields in which technology is not a key characteristic: for example in the field of arts, Lecoutre et al. \cite{Lecoutre2017RecognizingLearning}, have used a \textit{residual neural network} (ResNet) \cite{He2016DeepRecognition} to build a model capable of detecting the artistic style of a painting with an accuracy of 62$\%$., which could help in future the indexing of art collections.

Despite the vast range of applications, one industry in which the benefit of these tools has yet to be seen in their full potential is the Construction industry. This industry is known for its slow rate of improvement compared to the majority of other industries. As shown in \cite{Fulford2014ConstructionPractice}, one of the main reason for this is the lack of investment in technology, contrary to many other industrial fields. Another reason for the lack of substantial machine learning applications in the construction business, is the poor availability of data, essential for any machine learning tool. 

Nowadays, it has become widely common the use of \textit{Building Information Modelling} (BIM). The BIM is a tool which allows to maintain a digital representation of the building information in all its aspects \cite{Gu2010UnderstandingIndustry}, therefore allowing to store virtually all data related to a specific structure, concerning both its geometric as well as non-geometric aspects.
\begin{figure}[t]
\centerline{\includegraphics[scale=0.8]{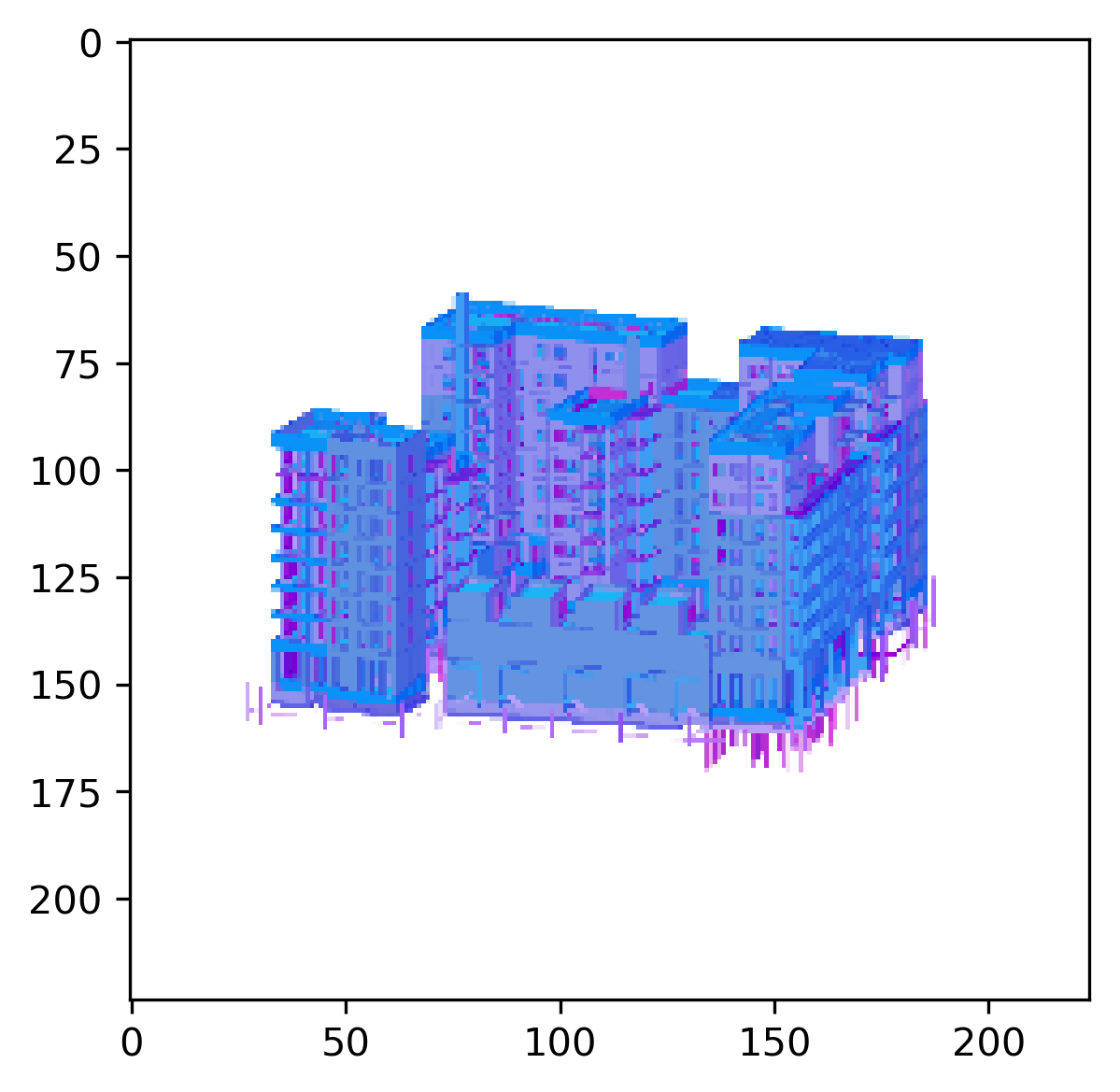}}
\caption{Example of a BIM virtual representation of an apartment building as presented to the models. The image has a dimension of 224x224 pixels.}
\label{fig_1}
\end{figure}

Although the BIM designs are stored in a digital format, there is indeed a need for automating their categorization: showing that it is possible to automatically categorize a structure type from a BIM representation, this approach could be extended to recognize more specific details or areas within the same models. Moreover, it could be possible to categorize the thousands of models which are already available in the databases, without the need for manually going over each and every of them. In this way, the presence of the historical data combined with the acquired knowledge of the building type's key features could help in the future when developing the methods to automatically design structures.

In this research, some machine learning models, more specifically related to image recognition task, have been used to automatically recognize the type of the building based on its BIM virtual representation. With this methods it was possible to separate BIM data in three different categories: apartment building, industrial building and others. As the name of the categories suggest, the first class includes images related to apartment buildings, the second includes images related to industrial buildings, and the class \textit{other} includes images of building which don't belong to neither of the former two. Other studies have applied deep learning to recognize building types, but only using real life images \cite{Bezak2016BuildingLearning}. The objective of this work is to show that it is possible to recognize structure types even when an image of the real structure is unavailable, training the models using only images generated from BIM structures.

The approach used focuses on four different machine learning models: the first model is based on the use of a Support Vector Machine (SVM)\cite{Cortes1995Support-vectorNetworks} to classify images, whose features have been previously extracted using Histograms of Oriented Gradients (HOG)\cite{Dalal2005HistogramsDetection}. The other three models are instead based on deep learning: the first two use \textit{pre-trained} \textit{Convolutional Neural Network} (CNN) architecture, more specifically a MobileNet \cite{HowardMobileNets:Applications} and a Residual Network (ResNet) \cite{He2016DeepRecognition}. These network have been pre-trained on the ImageNet dataset\footnote{http://image-net.org/} \cite{JiaDeng2009ImageNet:Database}, a collection of real life images classified in several thousands classes. The last model is a CNN with a randomly generated structure. The models' accuracies have been validated on a database of 240 images of structures (Figure \ref{fig_1}) extracted from 60 BIM virtual representations (4 images extracted for each BIM).

\begin{figure}[!t]
\begin{center}
	\subfloat{\fbox{\includegraphics[scale=0.6]{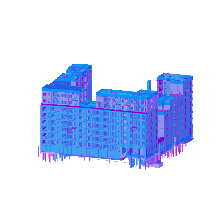}}}
    \subfloat{\fbox{\includegraphics[scale=0.6]{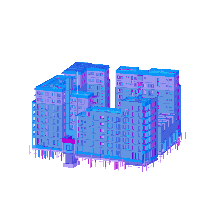}}}
    
    \squeezeup
    \subfloat{\fbox{\includegraphics[scale=0.6]{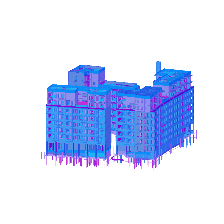}}}
    \subfloat{\fbox{\includegraphics[scale=0.6]{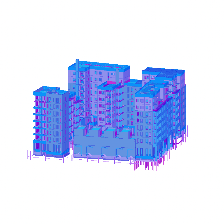}}}
    \caption{Example of the different images for each BIM structure: each image is generate from a completely different angle of the structure.}
    \label{angles_example}
\end{center}   
\end{figure}

The choice of these models is related to the dataset presented: having a dataset of thousands of BIM structures, but only a portion of them labeled, deep learning was chosen as it is easier to scale to bigger number of samples compared to classical machine learning. For this reason the neural networks were chosen. They all present a lightweight structure (particularly the MobileNet) and maintain a low level of complexity (remarkably in the case of the ResNet); moreover the random generated structure present a much shallower architecture and allows for an even lower computational complexity. The support vector machine is presented to compare the performance of the neural networks with a classical machine learning model.

This paper is structured as follow: Section II will describe the data and the methods used, Section III will show the results obtained with the experiments and comment them, and Section IV will discuss the conclusion and future work. 

\begin{figure*}[!t]
	\subfloat{\fbox{\includegraphics[scale=0.42]{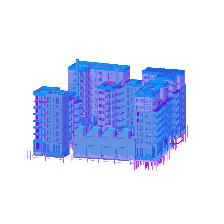}}}
    \subfloat{\fbox{\includegraphics[scale=0.42]{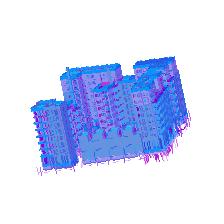}}}
    \subfloat{\fbox{\includegraphics[scale=0.42]{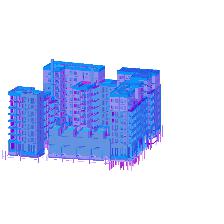}}}
    \subfloat{\fbox{\includegraphics[scale=0.42]{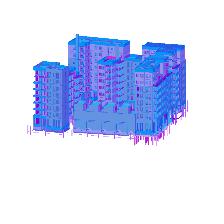}}}
    \subfloat{\fbox{\includegraphics[scale=0.42]{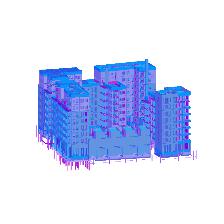}}}
    \caption{Example of augmented image as shown to the network. From left to right it can be seen: the original image, a rotation, horizontal shift, vertical shift and a horizontal flip.}
    \label{figura2}
\end{figure*}

\section{Methods}
Following it will be describe the four different machine learning models proposed for the building type recognition. The first is based on a classical machine learning approach, using Histograms of Oriented Gradients (HOG) for feature extraction and a Support Vector Machine (SVM) for classification, while the remaining three consist of two convolutional neural network (CNN) pre-trained on ImageNet dataset, and one CNN generated with random structure and random initialized weights (not pre-trained).

Prior to describing the models used, the data used will be described.

\subsection{Data}

The database used in this work was collected with the help of Sweco Structures Ltd. It consists of a total of 240 structural models, in which the images were extracted from their BIM virtual representations: 4 images for each of the 60 BIM representation have been extracted, showing completely different angles of the structure (Figure \ref{angles_example}). 

Due to the low number of images available to validate the deep learning models, they have been augmented randomly generating samples, processed with a combination of random rotations, horizontal flips and vertical and horizontal shifts (Figure \ref{figura2}). 

\begin{figure}[!ht]
	\subfloat{\fbox{\includegraphics[scale=0.33]{original.jpeg}}}
    \subfloat{\fbox{\includegraphics[scale=0.33]{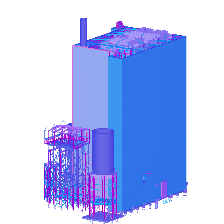}}}
    \subfloat{\fbox{\includegraphics[scale=0.33]{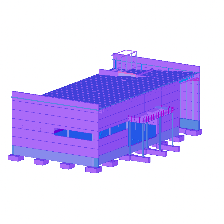}}}
    \caption{Example images for each class. From left to right: an apartment building, an industrial building and a structure belonging to the class "Other".}
    \label{classes_examples}
\end{figure}

The images are assigned in 3 classes: Apartment building, Industrial building and Other (Figure \ref{classes_examples}).

In the latter a variety of building not belonging to neither of the first two classes is present.
The images have been scaled to a dimension of 224x224 pixels to match the requirement of the two pre-trained neural networks.

\subsection{HOG + SVM}
The classical machine learning method chosen for the problem described, is a combination of Histograms of Oriented Gradients (HOG) used for feature extraction and Support Vector Machine (SVM)\cite{Cortes1995Support-vectorNetworks} for the classification task.

\begin{figure}[H]
\begin{center}
	\subfloat{\includegraphics[scale=0.45]{original.jpeg}}
    \subfloat{\includegraphics[scale=0.36]{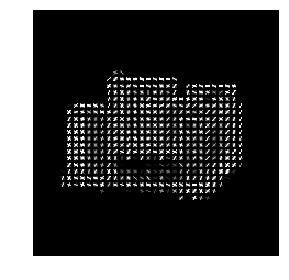}}
	\caption{Example of histograms of oriented gradients computed for one of the images in the dataset.}
	\label{hog}
\end{center}
\end{figure}

As described in \cite{Dalal2005HistogramsDetection}, by dividing the image in \textit{cells} and calculating for each pixel in the cell a local 1-D histogram of gradient direction, an object in the image can be individuated even without knowing the exact position of the gradient: the magnitude of the histograms of oriented gradients is larger for the edges of the objects. The gradients are then accumulate over larger \textit{blocks}, creating a dense grid of HOG \textit{descriptors}. In Figure \ref{hog} it can be seen an example of HOG descriptor for one of the images of the dataset.

For the problem described in this paper, the best performing set of parameter was found to be 4x4 pixel cells and 2x2 cell blocks. Once the histograms of gradient are computed, a feature descriptor is created for each image. The descriptors are used to train a multiclass SVM with linear kernel, which is used to classify the structure type.

\subsection{Pre-trained Neural Network}

In the past years, huge progress has been made in computer vision with the use of deep neural network models. Thanks to the availability of increasingly powerful computation hardware, it has been possible to develop and test very deep neural network architectures.

Amongst the methods to evaluate how a deep neural network perform in computer vision, a widely used benchmark is given by the ImageNet database \cite{JiaDeng2009ImageNet:Database}, a collection of real life images categorized in several thousands classes.

For this work, even if the database used includes only images extracted from virtual models, we used two of the top performing deep neural networks architectures on the ImageNet database: the MobileNet \cite{HowardMobileNets:Applications} and the ResNet \cite{He2016DeepRecognition}. 

\subsubsection{MobileNet}
Released by the Google research team, this model is optimize to work on mobile or embedded vision applications. Contrary to the traditional CNN, the MobileNet factorize the standard convolution performed in the convolutional layers by a CNN, into a depthwise convolution and a 1x1 pointwise convolution. This allows the model to first filter and in a second step combine the inputs to a new set of outputs, using two different layers. In this way both the computation and the model size are reduced \cite{HowardMobileNets:Applications}.

Another features of this architecture is the presence of two additional hyperparameters: the \textit{width multiplier} and the \textit{resolution multiplier}. The first, indicated with $\alpha$, is used to uniformly reduce the number of input and output channels, reducing the number of parameters and therefore the computational cost. The second, indicated with $\rho$, reduces the image representation throughout the network.

In this paper, the parameter $\alpha$ was assigned the values of 0.25, 0.5, 0.75 and 1, and the model was validated for each of them.

In addition, the last Dense layer of the MobileNet was substituted with one which would fit this research needs: this layer maps the output of the network to the three classes in which the images will be categorized.

\begin{figure*}[t]
\centerline{\includegraphics[scale=0.36]{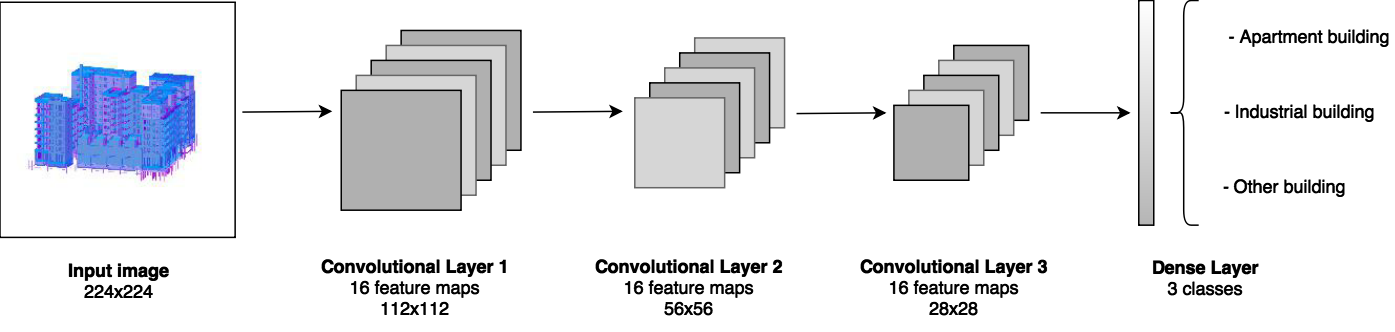}}
\caption{Structure of the neural network with random generated structure: it presents 3 convolutional layers, followed by a dense layers that maps the output of the last convolutional layer to the three classes. }
\label{fig_3}
\end{figure*}

\subsubsection{ResNet}
The ResNet \cite{He2016DeepRecognition} was released by the Microsoft research team and was the winner of the ImageNet Large Scale Visual Recognition Challenge (ILSVRC) in 2015\footnote{http://www.image-net.org/challenges/LSVRC/2015/results}.

The peculiarity of this particular type of CNN, is that it introduces the concept of \textit{residual learning}. The use of deep network architecture for image recognition has allowed a series of breakthroughs, and therefore there has been the tendency to create very deep neural network. Unfortunately, the deeper a neural network, the more difficult becomes the training as the accuracy tends to saturate. Through the use of residual learning, the network can learn \textit{residuals}, in other words the network presents \textit{shortcut connection}, which directly connect a layer $n$ to the layer $(X+n)$. \textit{He et al.} proved that these types of networks are easier to optimize and can gain accuracy from an increased depth, performing better than shallower networks, because of their ability to maintain a lower complexity.

For this research it was used a ResNet50, a residual network with 50 layers, whose weights were pre-trained on the ImageNet database.
As for the MobileNet, the last dense layer of the network was substituted with one which maps the output of the network into the three classes defined before

\begin{table*}[!t]
\renewcommand{\arraystretch}{1.4}
\caption{List of hyper-parameters selected through random search: in the middle column the range of values to be randomly selected, and in the right column the best choice}
\label{hyperparam}
\centering
\begin{tabular}{l|c|c}
\hline
\bfseries Hyper-parameter & \bfseries Value range & \bfseries Random Search Choice\\
\hline\hline
Number of Convolutional Layers & 1 -- 5 & 3\\
Number of Maps & (16, 32, 48) & 16\\
Kernel Size & 1 -- 5 & 2\\
Batch Normalization Layers & (0, 1) & 0\\
Dropout Rate & (0.2, 0.3, 0.4, 0.5) & 0.3\\
Learning Rate & $10^{-5}$ -- $10^{-1}$  & 0.008130275
\end{tabular}
\end{table*}

\subsection{Neural Network with Random Structure}
In addition to the two pre-trained network just described, it was also used a CNN with a randomly generated structure and randomly initialized weights (not pre-trained). The choice to randomly generate the network structure is linked to one of the problems in building efficient neural network architecture: it is often difficult to optimize and fine tune its parameters. There are different methods to optimize a network parameters, one of which is to randomly generate its structure. In \cite{Bergstra2012RandomOptimization}, the authors have proven that using randomly chosen hyper-parameter combination to optimize a model, is more efficient than manually searching for them or using a grid search: with a random search approach to optimize a neural network hyper-parameters, it is possible to deliver better models, while utilizing a much smaller computation time.

For this reason it was used a random search algorithm to generate the CNN structure. In Table \ref{hyperparam}, it is shown the hyper-parameters that were optimized using a random search algorithm. In particular it can be seen that for this specific case the network presents between 1 and 5 \textit{Convolutional layers}, each with the same number of \textit{Feature maps}, comprised between 16 and 48, and with \textit{Kernel size} between 1 and 5. It was also randomly chosen the \textit{Dropout rate}, which has a value between 0.2 and 0.5, and it is a parameter used to prevent the network from overfitting, and more importantly the \textit{Learning rate}, randomly chosen between $10^{-5}$ and $10^{-1}$. Moreover, there is the possibility to have a Batch Normalization layer, a layer which perform the normalization for each training batch, allowing the use of higher learning rates while still achieving high values of accuracy, with fewer iterations \cite{Ioffe2015BatchShift}.

To find the best structure for the neural network, the random search algorithm builds various models with random combinations of the hyper-parameters described above and validates each model with a 3-fold cross validation. The best model is chosen as the one that scored the highest validation accuracy, which is calculated as the mean of the validation accuracies of each of the 3-fold evaluation.

The best performing structure for this specific case is shown in Figure \ref{fig_3}: it includes $3$ convolutional layers, each with $16$ feature maps and a kernel size of $2$. Moreover it doesn't present a batch normalization layer. Finally the structure presents a dropout rate of $0.3$. The resulting neural network is optimized with a \textit{Stochastic Gradient Descent} which presents a learning rate of $0.008130275$.

Each convolutional layer uses a \textit{rectified linear unit} (ReLU) activation function \cite{Glorot2011DeepNetworks}, and the feature maps produced by each convolutional layer are downsampled by a factor of 2 with a max-pooling. 
The feature maps of the last convolutional layer are \textit{flatten}, and then mapped in the last layer, a dense layer, to the three classes defined before. 

\begin{table*}[!t]
\renewcommand{\arraystretch}{1.4}
\caption{Accuracy result on the validation set (which consists of $20\%$ of the dataset described before)}
\label{accuracy}
\centering
\begin{tabular}{l|c}
\hline
\bfseries Model & \bfseries Accuracy \\
\hline\hline
HOG + SVM & $57.19\% \pm 1.18\%$ \\
\hline
ResNet50 & $\bf 97.92\% \pm 1.32\%$\\
\hline
MobileNet $\alpha = 1$ & $93.75\% \pm 2.94\%$\\
MobileNet $\alpha = 0.75$ & $95.42\% \pm 2.75\%$\\
MobileNet $\alpha = 0.5$ & $94.62\% \pm 3.35\%$ \\
MobileNet $\alpha = 0.25$ & $91.70\% \pm 2.58\%$ \\
\hline
CNN with randomly generated structure & $89.60\% \pm 3.39\%$ \\
\end{tabular}
\end{table*}

\section{Results}

All the neural network used for this research have been implemented in Python, using Keras API \cite{chollet2015keras}. The Support Vector Machine has been implemented using Scikit-Learn \cite{Pedregosa2012Scikit-learn:Python}. Scikit-Image\cite{vanderWalt2014Scikit-image:Python} has been used to compute the Histograms of Oriented Gradients.
To evaluate their accuracy accuracy, the database of 240 images described before was randomly split in order to use $80\%$ of the database to train the models and the remaining $20\%$ to validate them. All the models were validate using a 5-fold cross-validation: the accuracy for the models was obtained calculating the mean of the 5-folds accuracies alongside its standard deviation.

In Table \ref{accuracy}, the results of the best model evaluation are shown: the best performing neural network was the ResNet50, with an accuracy of $97.92\% \pm 1.32\%$. The MobileNet as well scored an accuracy above $90\%$, in particular the MobileNet with $\alpha=1$ (the baseline network presented in \cite{HowardMobileNets:Applications}), scored an accuracy of $93.75\% \pm 2.94\%$. For the other settings of the $\alpha$ parameter, the accuracy obtained was best for $\alpha = 0.75$ ($95.42\% \pm 2.75\%$) followed by the reduced MobileNets with $\alpha = 0.5$ and $\alpha = 0.25$ (respectively of $94.62\% \pm 3.35\%$ and $91.70\% \pm 2.58\%$).
Moreover, it can be seen that the CNN with randomly generated structure performed well, obtaining an accuracy of $89.60\% \pm 3.39\%$, lower than the ones obtained with the pre-trained network, but still acceptable.

The worst performing model was the HOG + SVM approach, which scored an accuracy of only $57.19\% \pm 1.18\%$, much lower than the worst performing neural network. This indicates that a shallow model bases its prediction on local features and is unable to model the global structure. Our data contains several repetitive patterns that are not directly to the category, which makes the classical HoG representation unsuitable for this case. 

Moreover, these results show that despite the small dimension of the dataset used, the deep learning models outperformed the classical machine learning model, therefore supporting our thesis to use a deep learning approach also for future analysis.
It is also important to notice that although the dataset included only 240 images, by randomly augmenting them it was possible to show enough samples to the network in order for it to properly learn the key features of the image of the structures, without incurring in problems related to overtraining.

\begin{figure}[!b]
\begin{center}
	\subfloat{\fbox{\includegraphics[scale=0.6]{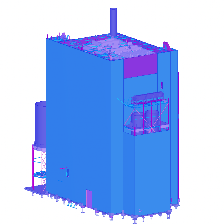}}}
    \subfloat{\fbox{\includegraphics[scale=0.6]{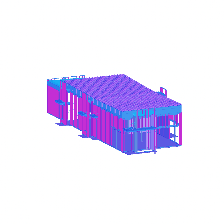}}}
	\caption{Example of misclassified images: the image on the left was categorized as \textit{Other}, while the correct class was \textit{Industrial building}; the image on the right was categorized as \textit{Apartment building} while the correct class was \textit{Other}.}
	\label{wrong}
\end{center}
\end{figure}
As an example of which type of images were misclassified, in Figure \ref{wrong} two examples are presented. The picture on the left, whose correct category is Industrial building, was misclassified as Other: it can be seen that also to a human it might resemble a warehouse, one of the building category that was included in the class Other. Similarly, the image on the right, was misclassified as an Apartment building, while in reality it belongs to the class Other. The errors committed by the neural networks were related to this latter class: in fact, it includes a range of different type of structures which weren't numerous enough to be allocated in stand-alone classes.

\section{Conclusion}

In this paper it was shown an application of image recognition to an artificial image. More specifically, the structures' images used were extracted from BIM program, and successfully categorized in three different classes: Apartment building, Industrial building and Other building types.

	Moreover it was shown that for this type of classification based on artificial images, it is possible to obtain good accuracy results both using neural networks pre-trained on real life images (MobilNet and ResNet) and neural networks with randomly generated structure and randomly initialized weights. This will allow to scale these models for the classification of the thousands of BIM structures which are yet to be labeled.
    
Furthermore, we believe that this type of work could be seen as one step towards the automation of the buildings' designing process: it shows that it is possible to successfully recognize the building type from an image extracted from a virtual model. Moreover, this approach could be extended to more specific details within the same model. In this way the machine learning algorithm could learn the key features of a building belonging to a certain category, and this acquired knowledge could be used in the future when designing the methods to automatically design other structure based on BIM historical data.

It is important, however, to notice that the dataset used presented very few images (240), and therefore it was necessary to randomly augment them in order to show enough sample to the neural network and obtain reasonable results while avoiding overtraining. It is also important to notice that for this work we only used 3 classes: as we mentioned before, we weren't able to create other stand-alone classes due to the few number of samples readily available for each category other than "Apartment building" and "Industrial building". For this reason we decided to put them together under the same label "Others".

For the reasons explained, our future plan is to label more samples in order to build a substantial dataset and simultaneously extend the number of classes considered. We also plan to extend this image recognition methods to further subdivide the BIM main categories into sub-categories that could represent different areas of interest in these structures.

\section*{Acknowledgment}

This research was funded by KIRA-digi project and Sweco Structures Ltd.

The authors would also like to thank Sweco Structures Ltd. for providing the dataset used for this work as well as for the valuable knowledge of the construction industry.



\end{document}